%% file: root.tex
\documentclass[letterpaper, 10 pt, conference]{ieeeconf}  

\IEEEoverridecommandlockouts                              

\usepackage{graphicx} 

\usepackage{color}
\usepackage{float}
\usepackage[font=scriptsize]{subfig}

\usepackage{algorithm}
\usepackage{algpseudocode}
\usepackage{tikz} 
\usepackage{hyperref}

\graphicspath{{figures/}}

\title{\LARGE \bf \center
Grounding of the Functional Object-Oriented Network \newline in Industrial Tasks*
}

\author{Rafik Ayari, Matteo Pantano\authorrefmark{2} and David Paulius
\thanks{*This work was not supported by any organization}

\thanks{Rafik Ayari, Matteo Pantano, and David Paulius are affiliated with the Human-Centered Assistive Robotics (HCR), Department of Electrical and Computer Engineering, Technical University of Munich, Munich 80333, Germany.}
\thanks{Matteo Pantano is also with Functional Materials \& Manufacturing Processes, Technology Department, Siemens Aktiengesellschaft, Munich 81739, Germany.}
\thanks{\authorrefmark{2}~Corresponding author: {\tt matteo.pantano@siemens.com}}
}

\begin{document}

\maketitle

\thispagestyle{empty}
\pagestyle{empty}

\begin{abstract}

In this preliminary work, we propose to design an activity recognition system that is suitable for Industrie 4.0 (I4.0) applications, especially focusing on Learning from Demonstration (LfD) in collaborative robot tasks. More precisely, we focus on the issue of data exchange between an activity recognition system and a collaborative robotic system. We propose an activity recognition system with linked data using \textit{functional object-oriented network} (FOON) to facilitate industrial use cases. Initially, we drafted a FOON for our use case. Afterwards, an action is estimated by using object and hand recognition systems coupled with a recurrent neural network, which refers to FOON objects and states. Finally, the detected action is shared via a context broker using an existing linked data model, thus enabling the robotic system to interpret the action and execute it afterwards. Our initial results show that FOON can be used for an industrial use case and that we can use existing linked data models in LfD applications.


\end{abstract}

\section{INTRODUCTION}

With the a priori defined 4\textsuperscript{th} Industrial revolution (I4.0) the concept of connected factories is de facto standard for competitive advantages of European manufacturing \cite{Forschungsunion.2013, Hermann.2016}. The concept envisages that the Factories of the Future (FoF) will be able to share information regarding processes and products that will enable life-cycle tracking of development, production, and services \cite{DeutschesInstitutfurNormung.2016.04}. However, the realization of this vision is still far from reaching modern companies \cite{Legat.2014}. Automation will play an important role in the domain of FoF, and the National Institute of Standards and Technology (NIST) has outlined the main challenges in the manufacturing interoperability program of NIST \cite{Kemmerer.2009}. Among those, the most important to our work is the misinterpretation of definitions or meaning of terms when exchanging information, which is mainly due to the lack of common data models that can be easily shared among factories \cite{Legat.2014}. Therefore, in this work, we would like to address this issue by analysing an I4.0 related industrial use case where a collaborative robot system needs to communicate with an activity recognition system to learn about a new task using Learning from Demonstration (LfD). To accomplish this, we will use the \textit{functional object-oriented network} (abbreviated as FOON)~\cite{Paulius2016FOON,Paulius2018FOON,Paulius2021FOON} together with a semantically linked data model.

The paper is organised as follows. In Section~\ref{sec:background}, we describe prior work in this area and discuss the need for a semantically linked data model to improve interoperability considering a use case in LfD. Afterwards, in Section~\ref{sec:usecaseandmethodology}, we explain the addressed use case and the methodology used for the action recognition together with its grounded FOON data model. Finally, in Section~\ref{sec:workinprogress}, we discuss our preliminary results and the future steps of our research.




\section{BACKGROUND \label{sec:background}}

\subsection{Functional Object-Oriented Network}
FOON is a symbolic knowledge representation for robots that takes the form of a semantic graph.
Motivated by the theory of affordance~\cite{Gibson_1977}, a FOON can be used to describe objects, actions, and their corresponding effects on those objects or on the robot's environment.
This is achieved through its bipartite network structure comprising of two types of nodes: object nodes and motion nodes.
Object nodes describe any sort of object that is used in manipulation or activities, such as tools, ingredients or components, while motion nodes describe actions using human action labels, which can be either manipulation actions (e.g., pouring or cutting) or non-manipulation actions (e.g., baking or frying).

\begin{figure}
    \centering
    \subfloat[Assembly FOON]{\includegraphics[width=\columnwidth]{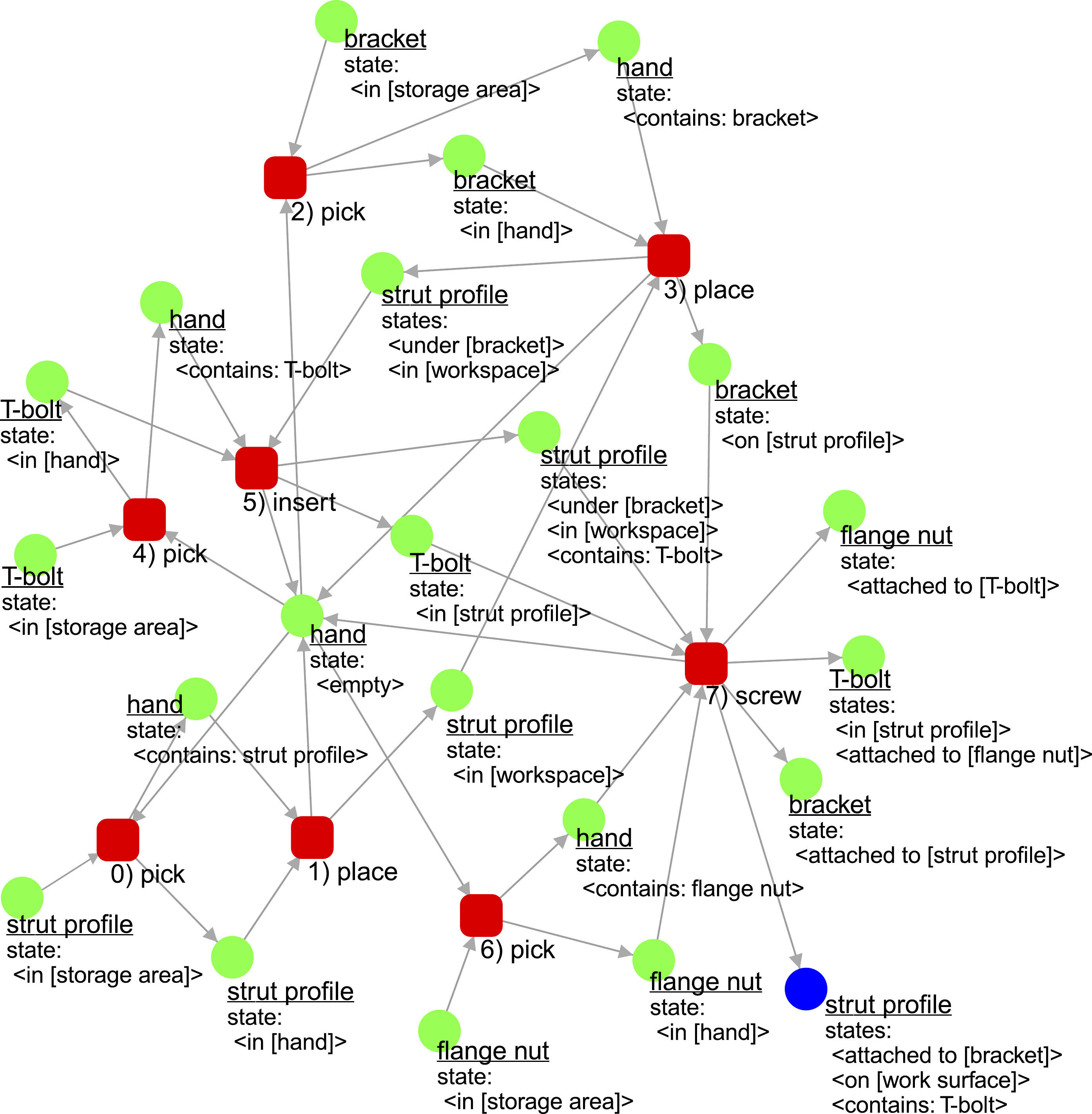}\label{fig:FOON-ass}}
    
    \subfloat[Disassembly FOON]{\includegraphics[width=\columnwidth]{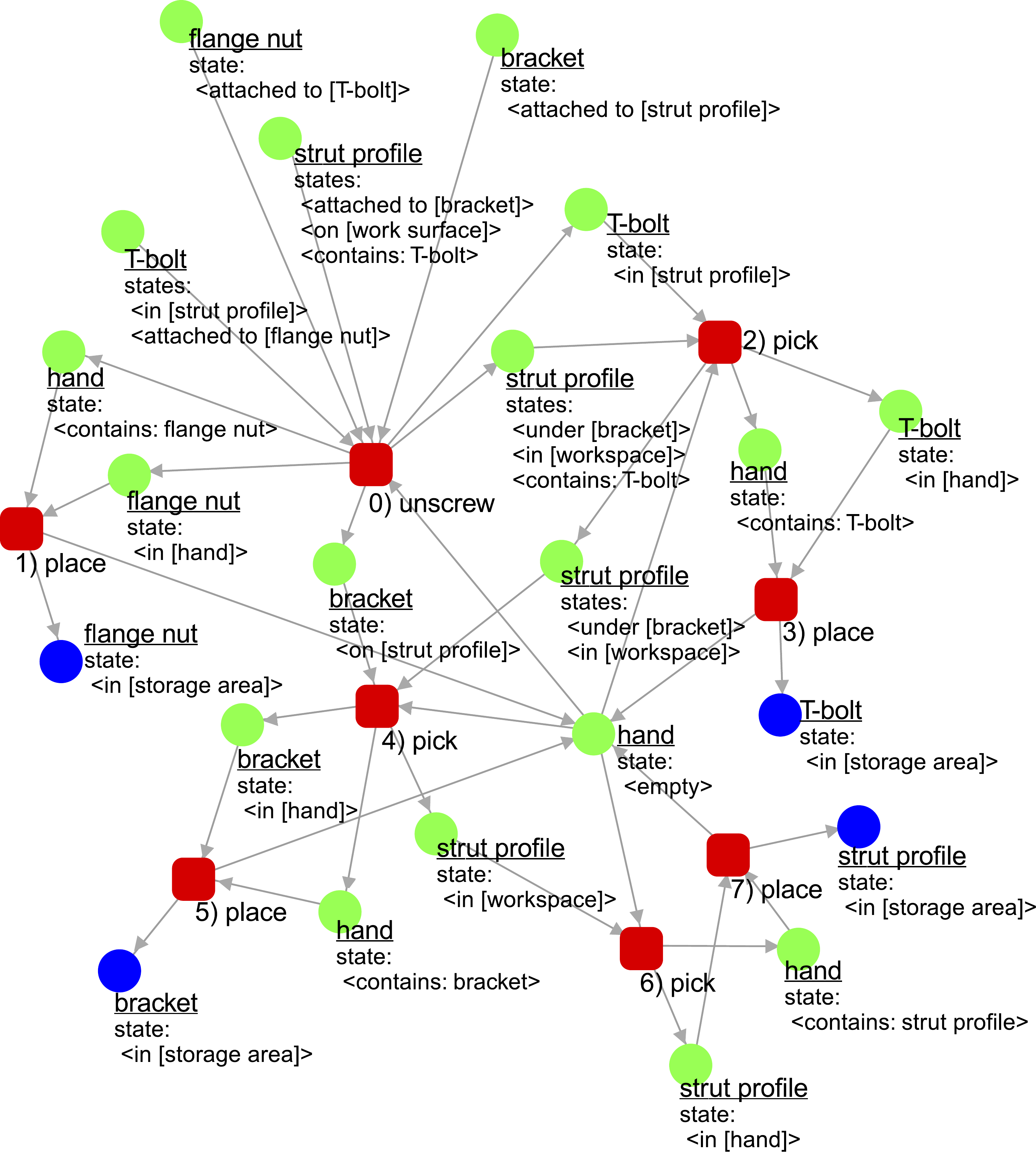}    \label{fig:FOON-dis}}
    \caption{Illustrations of FOON subgraphs for: a) industrial assembly and b) disassembly of parts (best viewed in colour). Here, object nodes are denoted by circles, while motion nodes are denoted by squares. Regular object nodes are denoted in lime green, while intended goal nodes for each subgraph's task is denoted in blue. For the assembly task, the goal is to attach a bracket to a strut profile and secure it with a T-bolt and flange nut, while for the disassembly task, the goal is to dismantle the secured bracket and strut profile. 
    }
    \label{fig:FOON}
\end{figure}

As a bipartite graph, object nodes are connected to motion nodes, while motion nodes are connected to object nodes. 
This innately allows a FOON graph to describe an object's state change(s) as actions are executed, as a motion node's incoming edge indicates a precondition, while a motion node's outgoing edge indicates an effect.
In FOON terminology, an action is described as a \textit{functional unit}, which comprises of input and output objects (describing precondition and effects) and a motion node.
A chain of functional units describing a single activity is referred to as a \textit{FOON subgraph} (example shown in Figure~\ref{fig:FOON}), and a FOON graph that is made from two or more subgraphs is referred to as a \textit{universal FOON}.

\subsection{Interoperability for the Factories of the Future} 
As previously mentioned, interoperability is the field addressing the perfect exchange of information between I4.0 actors (e.g., machine, human), where perfect means that accurate information exchange is guaranteed during the data life-cycle (i.e., origin, production, and use) \cite{Ford.2007}. To address this issue in the manufacturing domain, the so-called reference architectures (RAs) have been introduced by several standardization bodies. RAs are blueprints that contain and track all the information regarding processes' life-cycles, thereby enabling faster reuse of information \cite{Han.2020}. The most known RAs are: Reference Architecture Model Industrie 4.0 (RAMI4.0) \cite{DeutschesInstitutfurNormung.2016.04}, ISO/ TC 184 \cite{InternationalOrganizationforStandardization.2018}, and NIST Smart Manufacturing Architecture \cite{Lu.2016}. However, when considering real manufacturing use cases, the RA with the most overlap with industrial applications is RAMI4.0 \cite{CruzSalazar.2019}. Of utmost importance in this RA are the Communication and Information layers, as they include the representation of data according to a uniform format \cite{DeutschesInstitutfurNormung.2016.04}. In these layers, the Open Platform Communications Unified Architecture (OPC UA) is often used. OPC UA communicates information using a shared semantic information model (i.e., ontology) while using a cross-platform transmission protocol \cite{StefanProfanter.2019}. Therefore, it could provide the information description as highlighted by the problem scope of this work. Unfortunately, the OPC UA information model\footnote{\url{https://opcfoundation.org/developer-tools/specifications-opc-ua-information-models}} does not have any activity description as it would be required for a LfD use case.

\subsection{Learning from Demonstration and Interoperability}
The joint design of an ontology and knowledge processing system has been a topic of interest for several domains and problems~\cite{nakawala2018approaches,manzoor2021ontology}.
One of the most well-known comprehensive knowledge processing systems and ontologies for service robotics is KnowRob~\cite{tenorth2013knowrob}, which is a part of the open-Ease project~\cite{beetz2015open} and was preceded by RoboEarth~\cite{waibel2011roboearth}. 
The motivation behind this ontology is to connect conceptual knowledge from several sources, such as natural language, online resources, and web pages, and to draw upon episodic memories and experiences from other robots as a cloud robotics platform. 
This work best demonstrates how an effective system that connects several modalities of data for robot task planning and execution can be designed. 
However, such a system consists of many components that are more suited for dynamic or diverse scenarios like in the household, and due to the unnecessary complexity, their advantages could not be fully integrated to industrial cases, especially when considering the relation between adoption and technology ease of use \cite{Saghafian.2021}.
Therefore, we choose to investigate a simpler system that will benefit from the intuitiveness of FOON and existing linked data modelling methods.



\section{USE CASE AND METHODOLOGY \label{sec:usecaseandmethodology}}

\subsection{Use Case}
\begin{figure}
    \centering
    \centering
    \def\svgwidth{\linewidth}
    \fontsize{11}{9} \selectfont
    \graphicspath{{figures/}} 
    \resizebox{0.92\linewidth}{!}{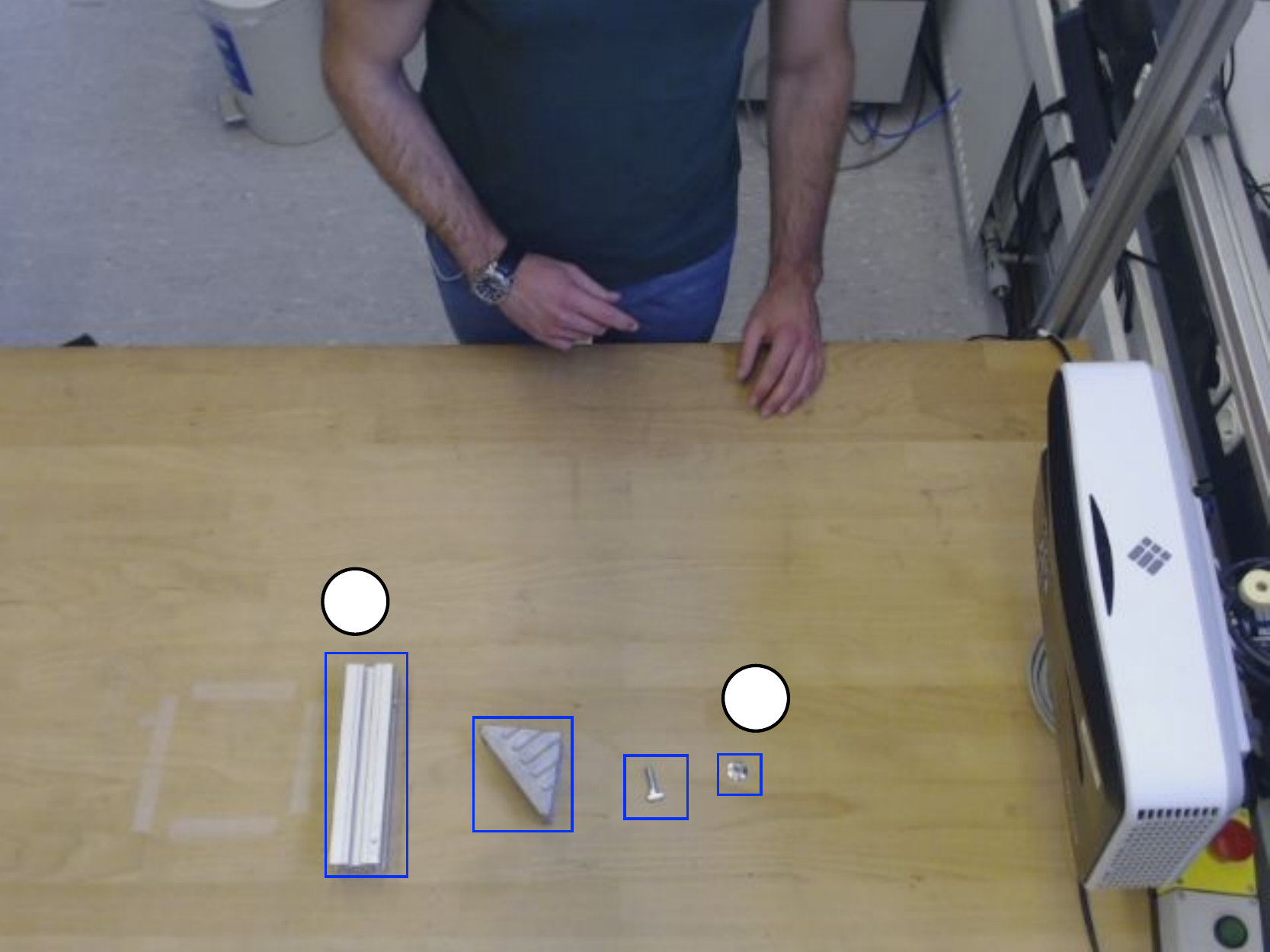}
    \caption{Image of the workbench with the operator and the parts used for the assembly process. Going from the left to the right, the parts are as follows: strut profile \raisebox{.2pt}{\textcircled{\raisebox{-.6pt} {1}}}, bracket \raisebox{.2pt}{\textcircled{\raisebox{-.6pt} {2}}}, T-bolt \raisebox{.2pt}{\textcircled{\raisebox{-.6pt} {3}}}, and flange nut \raisebox{.2pt}{\textcircled{\raisebox{-.6pt} {4}}}.}
    \label{fig:assembly}
\end{figure}

The use case analyses the assembly and disassembly of strut profiles. Strut profiles were chosen as they are conventionally used in the industry for building workcells (e.g., robotic workcell). 
The use case envisions an operator in front of a workbench with all the parts that needs to be manipulated as depicted in Figure~\ref{fig:assembly}. Considering the numbering in the Figure, the parts are: strut profile \raisebox{.2pt}{\textcircled{\raisebox{-.6pt} {1}}}, bracket \raisebox{.2pt}{\textcircled{\raisebox{-.6pt} {2}}}, T-bolt \raisebox{.2pt}{\textcircled{\raisebox{-.6pt} {3}}}, and flange nut \raisebox{.2pt}{\textcircled{\raisebox{-.6pt} {4}}}.

\subsection{Creating a FOON for Industrial Tasks}
FOON has been applied mainly to cooking activities. In this work, we apply the FOON representation to industrial tasks, specifically assembly and disassembly of parts required for preparing a robotic workcell.
Such parts include strut profiles, flange nuts, T-bolts, and brackets. 
For the time being, we constructed subgraphs for each task, and these are shown as Figures~\ref{fig:FOON-ass} and \ref{fig:FOON-dis} respectively.
These subgraphs describe several high-level skills or actions required to assemble components, such as picking-and-placing, screwing, and unscrewing. 
Skill definitions were taken from \cite{Backhaus.2017,Froschauer.2019}.
Furthermore, to make such graphs suitable for action recognition, the notion of hands or end-effectors were included to better identify objects of focus when using hand recognition, which will be addressed in Section~\ref{sec:workinprogress}.

\subsection{Modelling of the Ontology\label{subsec:modellingoftheontology}}
To address the identified need of an activity description aligned to an industrial-like infrastructure, we opted to use a semantically linked activity information model. Such decision was driven by the possibility of easily mapping a semantically linked model to the OPC UA information model as outlined by \cite{Schiekofer.2019}. Considering that we are using the FOON comprised of several \textit{functional units}, our modelling proposes an approach to build and connect \textit{functional units}. For our modelling, we start from the \textit{Task} data model\footnote{\url{https://shop4cf.github.io/data-models/task.html}} and \textit{Resources} data models\footnote{\url{https://shop4cf.github.io/data-models}} created by the Smart Human Oriented Platform for Connected Factories (SHOP4CF) consortium \cite{MichaZimniewicz.2020}. These information models were selected because they are based on linked data and they represent manufacturing use cases; therefore, they satisfy the requirements for mapping into OPC UA. The \textit{Task} data model was chosen due to its similar definition to a \textit{motion node}: a \textit{Task} is intended to be a manufacturing operation that may contain sub-steps, and it is associated to resources (i.e., persons, devices, materials, and assets) and locations. Moreover, the \textit{Resources} data models were chosen due to their definition of abstractly representing objects that are located on the shop floor. Accordingly, the mapping between \textit{functional unit}, \textit{Task}, and \textit{Resource} attributes can be summarized as follows. Input and output objects are mapped to the \textit{Resource} data models, as they can represent the \textit{state} of the objects (i.e., state change). Motion nodes are represented using the \textit{Task} data models, as they can specify which object is involved through the \textit{involves} property, and they can integrate the chain of events by using the \textit{isDefinedBy} property.

\section{WORK IN PROGRESS \label{sec:workinprogress}}

\subsection{Designing an Action Recognition Model}

\begin{figure}[!b]
    \centering
    \centering
    \def\svgwidth{\linewidth}
    \graphicspath{{figures/}} 
    \includegraphics[width=0.50\textwidth]{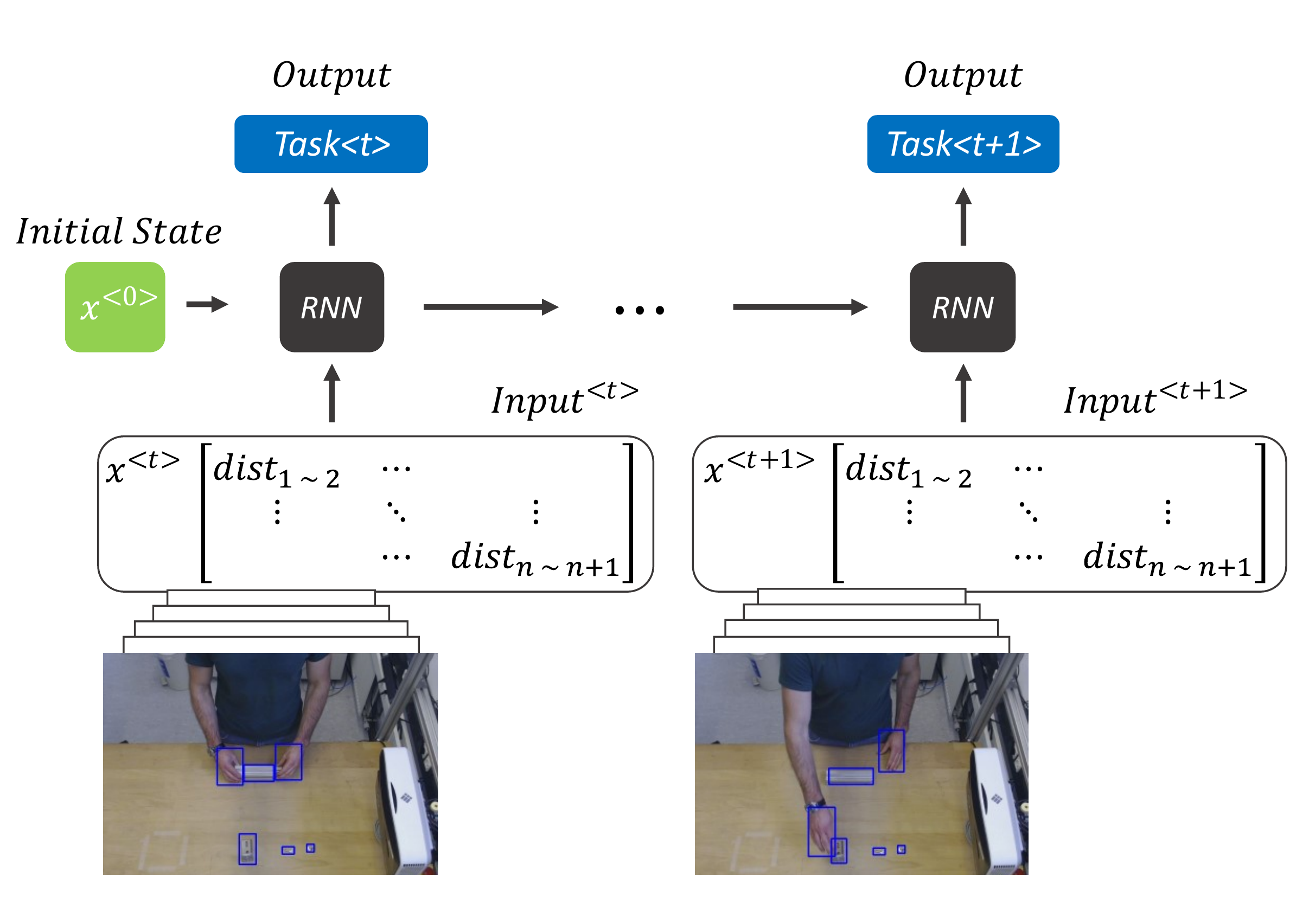}    
    \caption{Pipeline for the action recognition model. Initially, the classified images are scaled down, and the distances between object to object and object to hands are calculated and inserted into a matrix. Secondly, with the initial state of the scene, this matrix is fed to the recurrent neural network as input, which predicts the \textit{Task} associated with it.}
    \label{fig:pipeline}
\end{figure}

The aim of action recognition is to identify the action presently being performed by a human demonstrator or worker, which can be useful for collaborative robot (cobot) scenarios, where a robot needs to assist the human in completing a joint goal or task.
For the time being, we plan to employ a deep neural network framework to process hand-object information extracted by a YOLO-based~\cite{redmon2016you} pipeline to recognize actions as functional units in our FOON representation.
From a functional unit, we will know which objects are being used and in which state(s) they may currently be in as it relates to the task. Afterwards, with this input information in the format of an object-to-object distance matrix, object-to-hand distance matrix, and image, we plan to recognize the actions by using a recurrent neural network due to the sequential nature of the activities in our use case. The pipeline can be visualized in Figure~\ref{fig:pipeline}.
\newline

\subsection{Connecting FOON to an Ontology}
As stated in Section~\ref{subsec:modellingoftheontology}, we are using the linked information models \textit{Task} and \textit{Resources}. These models have been created using the NGSI-LD \cite{EuropeanTelecommunicationsStandardizationInstitute.2020} specifications and have been integrated within the FIWARE\footnote{\url{https://www.fiware.org/}} open-source platform. Therefore, the connection between the activity recognition and the information models will be integrated through this framework. Once the predicted unit is estimated via action recognition, the following information is added to the linked model. First, the unit is created by a new \textit{Task} model. Second, the objects related to the \textit{Task} are added to \textit{involves} property. Third, the \textit{Task} attribute \textit{isDefinedBy} is updated with a sequential ID referring to the modelled FOON above. Finally, the object state is updated according to the perceived state through the attribute \textit{state} and the \textit{Task} with \textit{Resources} are published on the FIWARE context broker. This approach is described by Algorithm \ref{alg:FOON2ont}.

\begin{algorithm}
		\caption{Connection of the FOON to the ontology}
		\label{alg:FOON2ont}
		\begin{algorithmic}[1]
			\Procedure{FOON2ont}{$PredictedUnit$}
			\State $Resources = []$
			\State $Task = []$
		    \For {each object $obj$ in $PredictedUnit$}
		        \State $Resource[state] = obj[state]$
		        \State $Resources.append(Resource)$
		        \State $Task[involves] = obj[id]$
		    \EndFor
		    \State $Task[isDefinedby] = PredictedUnit[id]$
			\State publish($Task$)
			\State publish($Resources$)
			\EndProcedure
		\end{algorithmic}
\end{algorithm}

\subsection{Future Work}
Once we have established a basic connection between the FOON representation and the ontology via FIWARE, we will further investigate ways to construct new graphs for other types of parts and unseen tasks. 
This can be achieved by treating object detection and motion recognition as separate pipelines. Moreover, a definition of the set of activities will follow in order to ensure a common representation of the set of activities both from the operator and the robotic system. This can be achieved by a common representation of the \textit{isDefinedBy}, which will point to a new entity \textit{Task Definition} with a pre-defined and explained vocabulary (e.g., JSON schema vocabulary\footnote{\url{https://json-schema.org/}}, SOMA ontology\footnote{\url{https://github.com/ease-crc/soma}}) of actions.


\bibliography{references}
\bibliographystyle{unsrt}

\end{document}

%% file: figures/parts.pdf_tex
\begingroup%
  \makeatletter%
  \providecommand\color[2][]{%
    \errmessage{(Inkscape) Color is used for the text in Inkscape, but the package 'color.sty' is not loaded}%
    \renewcommand\color[2][]{}%
  }%
  \providecommand\transparent[1]{%
    \errmessage{(Inkscape) Transparency is used (non-zero) for the text in Inkscape, but the package 'transparent.sty' is not loaded}%
    \renewcommand\transparent[1]{}%
  }%
  \providecommand\rotatebox[2]{#2}%
  \newcommand*\fsize{\dimexpr\f@size pt\relax}%
  \newcommand*\lineheight[1]{\fontsize{\fsize}{#1\fsize}\selectfont}%
  \ifx\svgwidth\undefined%
    \setlength{\unitlength}{479.99998558bp}%
    \ifx\svgscale\undefined%
      \relax%
    \else%
      \setlength{\unitlength}{\unitlength * \real{\svgscale}}%
    \fi%
  \else%
    \setlength{\unitlength}{\svgwidth}%
  \fi%
  \global\let\svgwidth\undefined%
  \global\let\svgscale\undefined%
  \makeatother%
  \begin{picture}(1,0.75000002)%
    \lineheight{1}%
    \setlength\tabcolsep{0pt}%
    \put(0,0){\includegraphics[width=\unitlength,page=1]{parts.pdf}}%
    \put(0.58317829,0.18602457){\makebox(0,0)[lt]{\lineheight{1.25}\smash{\begin{tabular}[t]{l}4\end{tabular}}}}%
    \put(0.26774243,0.26225953){\makebox(0,0)[lt]{\lineheight{1.25}\smash{\begin{tabular}[t]{l}1\end{tabular}}}}%
    \put(0,0){\includegraphics[width=\unitlength,page=2]{parts.pdf}}%
    \put(0.39866767,0.21751286){\makebox(0,0)[lt]{\lineheight{1.25}\smash{\begin{tabular}[t]{l}2\end{tabular}}}}%
    \put(0,0){\includegraphics[width=\unitlength,page=3]{parts.pdf}}%
    \put(0.50307635,0.19154874){\makebox(0,0)[lt]{\lineheight{1.25}\smash{\begin{tabular}[t]{l}3\end{tabular}}}}%
  \end{picture}%
\endgroup%